\documentclass[letterpaper, 10 pt, conference]{ieeeconf}  
\usepackage[utf8]{inputenc}
\pdfoutput=1

\IEEEoverridecommandlockouts                             
\usepackage{graphicx} 
\usepackage{amsmath} 
\usepackage{amssymb}  
\usepackage{color}
\usepackage{booktabs}
\usepackage{subcaption}
\usepackage{algorithm}
\usepackage{algpseudocode}
\usepackage{url}

\title{Vision-Based Road Detection using Contextual Blocks}

\author{Caio César Teodoro Mendes$^{1,2}$, Vincent Frémont$^2$ and Denis Fernando Wolf$^1$
\thanks{Authors$^1$ are with the Mobile Robotics Lab, 
Institute of Mathematics and Computer Science (ICMC), University of São Paulo (USP), 
SP, Brazil, {\tt\small \{caiom; denis\}@icmc.usp.br}}
\thanks{Authors$^2$ are with Heudiasyc UMR CNRS 7253, Université de Technologie de Compiègne, France, {\tt\small vincent.fremont@hds.utc.fr}}
}

\begin{document}

\maketitle
\thispagestyle{empty}
\pagestyle{empty}

\begin{abstract}
Road detection is a fundamental task in autonomous navigation systems. In this paper, we consider the case of monocular road detection, where images are segmented into road and non-road regions. Our starting point is the well-known machine learning approach, in which a classifier is trained to distinguish road and non-road regions based on hand-labeled images. We proceed by introducing the use of ``contextual blocks'' as an efficient way of providing contextual information to the classifier. Overall, the proposed methodology, including its image feature selection and classifier, was conceived with computational cost in mind, leaving room for optimized implementations. Regarding experiments, we perform a sensible evaluation of each phase and feature subset that composes our system. The results show a great benefit from using contextual blocks and demonstrate their computational efficiency. Finally, we submit our results to the KITTI road detection benchmark achieving scores comparable with state of the art methods.

\end{abstract}
\section{INTRODUCTION}

Autonomous vehicles, and more concretely Advanced Driver Assistance Systems (ADAS), can potentially reduce accidents, improve traffic flow, save fuel and consequently change the transport landscape. Road detection is a key component of such systems, providing not only free and valid space for maneuvers but also invaluable information for others tasks such as pedestrian and vehicle detection. 


In this work, we aim at estimating the road region using a monocular color camera. Visual road detection is a challenging task, where one has to deal with the continuously changing background, illumination issues, and most importantly, the high intra-class variability, i.e. the large variation in road appearance from place to place. Some works estimate the road area by relying on lane markings or sudden changes in appearance near the road boundaries. For instance, \cite{McCall:2006} uses steerable filters for robustly detecting lane markings. Another popular approach \cite{Passani:2014,vitor2014probabilistic} consist of using machine learning techniques, where a classifier is trained to distinguish between road and non-road regions based on images features (e.g. color and texture). In this context, many works focus on proposing new image features for road detection. In \cite{Alvarez:2011}, the authors propose the use of an illumination invariant color space to deal with shadowed areas.

A common limitation of most machine learning methods is that they independently classify each image region or pixel, ignoring the contextual information and are therefore subject to misclassifying areas of similar appearance. Some efforts have been made to address this issue; \cite{Kuhnl:2012} uses Conditional Random Fields (CRF) and \cite{Passani:2014} spatial rays features to incorporate contextual cues. Nevertheless they are limited because first-order CRFs only allow the direct influence of adjacent regions while spatial rays require a pre-segmented image. More powerful ways to exploit contextual information are presented in \cite{Mohan:2014} and \cite{Munoz:2010}, the former creates a hierarchical image segmentation, specific classifiers for each level of the hierarchic and uses the classification of one level as features to the next one. The later uses region-specific Convolutional Neural Networks (CNNs) allowing non-linear influence of distant regions. Both approaches, however, are computationally costly and were not able to reach real-time even with parallel implementations.

We hold that the key for reliable monocular road detection lies in the efficient use of contextual information, and consequently we propose a block scheme to efficiently incorporate contextual cues. Our method classifies small images patches using images features while the so-called ``contextual blocks'' provide contextual information. The other components, namely the image features and the classifier, were chosen taking into account not only their performance and adequateness to the task but also their computational cost, leaving room for optimized and possible real-time implementations.

The rest of the paper is organized as follows: Section \ref{sec:proposed_methodology} presents the methodology; Section \ref{sec:experiments} shows the method evaluation; The results are discussed in Section \ref{sec:discussion}; finally, Section \ref{sec:conclusions} draws the conclusions and suggests future works.

\section{Proposed Methodology}
\label{sec:proposed_methodology}

\begin{figure}
  \centering
    \includegraphics[width=0.48\textwidth]{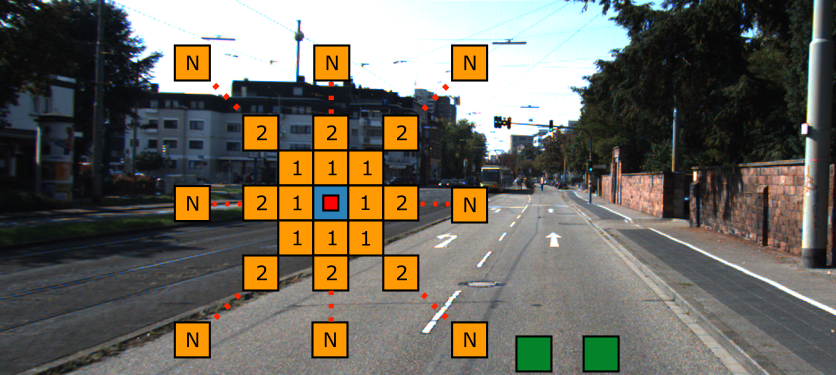}
      \caption{Illustration of the proposed block scheme. The classification block is show in red, the contextual blocks in orange, the possible support block in blue and the road blocks in green.}
 \label{fig:blockscheme}
\end{figure}

Our system makes extensive use of rectangular block/patches as shown in Fig. \ref{fig:blockscheme}. These blocks are divided into three categories: classifications blocks, contextual blocks and road blocks. Classification blocks are the ones whose pixels are classified while contextual and roads blocks are auxiliary and delimit regions from which features are extracted. To classify a single image region or classification block, one should extract features from the block itself, its respective contextual blocks and from the road blocks. All these features are pre-processed and concatenated into a single final vector $\mathbf{v}_{final} \in \mathbb{R}^N$ that is fed to a classifier. The output of the classifier, road or non-road, is attributed to every pixel of the classification block. To classify an entire image, this task should be repeated for every image region or classification block, and since we classify every pixel in a classification block, its vertical and horizontal stride is always equal to its vertical and horizontal size.

\begin{algorithm}[t]
\caption{Feature Concatenation}
\label{alg:feat}
\begin{algorithmic}[1]

\State $\mathbf{v}_{final} \gets \varnothing $\Comment{Empty vector}
\State $\mathbf{v}_{final} \gets \mathbf{v}_{final} \oplus \mathbf{v}_{class}$\Comment{$\oplus$: Concatenation}
\For {$i\gets 1$ \textbf{to} $radius\times8$}
    \State $\mathbf{v}_{final} \gets \mathbf{v}_{final} \oplus \mathbf{v}^i_{context}$
\EndFor
\If{$size(class) \neq size(context)$}
	\State $\mathbf{v}_{final} \gets \mathbf{v}_{final} \oplus \mathbf{v}_{support}$
\EndIf
\State $\mathbf{v}_{final} \gets \mathbf{v}_{final} \oplus (\mathbf{v}^1_{road} - \mathbf{v}_{class})$
\State $\mathbf{v}_{final} \gets \mathbf{v}_{final} \oplus (\mathbf{v}^2_{road} - \mathbf{v}_{class})$

\end{algorithmic}
\end{algorithm}

The features employed in this work do not provide spatial information, i.e. they do not make distinction between pixels positions within a block, hence we use what we call ``contextual blocks'' to provide information about the surroundings of the classification blocks. The first contextual blocks are positioned in the direct neighborhood of a reference block according to the eight connected scheme and further blocks are aligned in a ``star'' shape pattern. The reference block is the classification block itself if the classification block and the contextual blocks have the same size. Otherwise it consists of an additional block, called support block, centered on the classification block and with the same size as the contextual blocks. The number of contextual blocks for a classification one is given by their ``radius''. For instance, a radius of one yields 8 contextual blocks while a radius of 2 yields 16. The feature vector $\mathbf{v}^i_{context}$ of each one of the contextual blocks, and the possible support one $\mathbf{v}_{support}$, are concatenated into the final feature vector $\mathbf{v}_{final}$.

Finally, road blocks are positioned in the bottom part of the image and they provide a frame relative notion of the road appearance. The feature vector of each road block $\mathbf{v}^i_{road}$ is subtracted from the classification block feature vector $\mathbf{v}_{class}$ and concatenated into $\mathbf{v}_{final}$. The subtraction is made to directly provide the classifier a similarity notion of the block being classified and a supposed road region. We opted for using two small road blocks instead of a larger one, as it is usually done, to minimize the effect of lane markings in the road blocks features. The feature concatenation is summarized in Algorithm \ref{alg:feat}.

\subsection{Image Features}

In this paper, we decided to use some simple color and texture/structure features. We gave preference for fast (low computational cost) and low dimensional features. As small image regions are being classified, there is no need for complex features, such as those employed for object recognition (e.g. Histogram of Oriented Gradients). Furthermore, as we are using a parametric classifier, a low-dimensional feature vector is desirable since it can improve generalization.

Table \ref{tab:features} presents the selected image features. Entropy, Local Binary Patterns (LBP) and Leung-Malik (LM) \cite{Malik:2001} filters responses features are generated based on the grayscale image. The entropy is calculated using a circular support region with a radius of 5 pixels. For the LBP descriptor, we chose to use four neighbors instead of the usual eight reducing its histogram dimensionally from 256 to 16. We employed a subset of the original LM filter bank consisting of 6 edge, 6 bar, 1 Gaussian and 2 Laplacian of Gaussian filters, with a $19\times19$ pixel support, $\sqrt{2}$ scale for oriented and blob filters and 6 orientations.

\begin{table}[t]
\caption{Image features selection}
\centering
\begin{tabular}{@{}llc@{}}
\toprule
\multicolumn{2}{c}{Feature}                                & Dim. \\ \midrule
RGB          & Mean and std. dev. of each channel          & 6    \\
Grayscale    & Mean and std. dev.                          & 2    \\
Entropy      & Mean and std. dev.                          & 2    \\
LBP          & Normalized LBP histogram (4-connected)      & 16   \\
LM Filters 1 & Mean and std. dev. of filter responses      & 30   \\
LM Filters 2 & Normalized histogram of the max. responses  & 15   \\ \midrule
\multicolumn{2}{c}{Total}                                  & 71  \\ \midrule
\end{tabular}
\label{tab:features}
\end{table}

A spatial prior, in the form of the position of the classification block, is also included in our final feature vector. Preliminary tests suggested that it is preferred to input it encoded as a one-hot bit vector instead of a floating point. Intuitively, this encoding may facilitate the learning of strong priors in parametric models. Concretely we normalized each classification block coordinate, discretize it in 11 parts and represent each discretized coordinate as an 11 bins one-hot bit vector. Therefore the dimensionality of the spatial prior feature is $dim(\mathbf{v}_{spatial}) = 22$, $11$ for each coordinate. The exact number of bins should make a small difference in performance as long as it is not too small (e.g. $< 5$), compromising its discriminative power, or too large (e.g. $> 100$), significantly increasing the model complexity in parametric models.

If we assume the use of the additional support block, the dimensionality of the final feature vector is given by:
\begin{multline}
dim(\mathbf{v}_{final}) = dim(\mathbf{v}_{class}) + dim(\mathbf{v}_{support}) \\ 
\quad + radius \times 8 \times dim(\mathbf{v}_{context})\\
+ 2 \times dim(\mathbf{v}_{road}) + dim(\mathbf{v}_{spatial})
\end{multline}
where the function $dim$ returns the dimensionality of the input vector. It should be noted that, for this work $dim(\mathbf{v}_{class}) = dim(\mathbf{v}_{context}) = dim(\mathbf{v}_{road})$ and if we consider all features, they are all equal to 71.

\subsection{Classifier}

We chose to use a standard Multilayer Perceptron (MLP) neural network, which is a parametric non-linear model. MLPs present a reasonable classification performance in a wide range of tasks and are easily parallelizable to exploit the processing power of Graphics Processing Units (GPUs) and multi-core systems and, as it is a parametric model, its prediction computational cost does not depend on the training procedure (unlike SVMs, for instance).

Our model consists of one hidden layer with Rectified Linear (ReLU) activations functions and an output layer with the sigmoid activation function. We used the cross entropy cost function, therefore only one output neuron is used for the binary classification task. Formally, given that the feature vector $\mathbf{v}_{final}$ is a column vector the prediction is given by: 
\begin{equation}
g(\mathbf{v}_{final}) = \sigma(\mathbf{W}_{o} \cdot \psi(\mathbf{W}_h \cdot \mathbf{v}_{final})),
\end{equation}
where $\mathbf{W}_h$ and $\mathbf{W}_o$ are the weight matrices of the hidden and output layer respectively (each row stores the weights of a neuron), $\psi$ is the ReLU function and $\sigma$ is the sigmoid function. Finally, the output of the model is thresholded according to:

\begin{equation}
  L = \begin{cases}
   \text{Road} & \text{if} \quad g(\mathbf{v}_{final}) > 0.5 \\
   \text{Non-road} & \text{if} \quad g(\mathbf{v}_{final}) \leq 0.5 \\
  \end{cases}
 \label{eqjfunction},
\end{equation}
where $L$ is the label of the referent classification block.

For regularization, we limit the Euclidean norm of the MLP weights (parameters), the maximum value is chosen per layer and it is applied individually to the weights corresponding to a single neuron (output dimension of the layer). When the norm exceeds the limit, it is scaled down to have exactly the limit value. The training is done using mini-batch stochastic gradient descent with momentum. The training is finished after a number, here called of ``patience'', of epochs without any improvement in the accuracy of the validation set.

One drawback of MLPs is their large number of hyperparameters. To tackle this issue, before every training, we use a small subset of the training and validation sets to perform a hyperparameter optimization. For this optimization procedure we use the particle swarm optimization (PSO) algorithm and optimize the following parameters: number of neurons, learning rate, hidden layer maximum norm and output layer maximum norm.

\section{Experiments}
\label{sec:experiments}

\subsection{Dataset and Setup}

To evaluate our approach, we made use the KITTI Vision Benchmark Suite \cite{Fritsch:2013}. Specifically, we use the road detection benchmark, which provides 289 annotated images for training and 290 test images. Both sets are divided into three categories: urban unmarked (UU), urban marked (UM) and urban multiple marked lanes (UMM). Methods are ranked according to their pixel-wise maximum F-measure on the  Bird's-eye view (BEV) space. The benchmark further provides laser points (Velodyne data), stereo images and GPS data. In our work, only the monocular color images are used and we do not make distinction between the three road categories.

To evaluate each component of our system and to select the most adequate parameters/hyperparameters, we divide the 298 annotated images into a set of training/validation containing 260 images and a set for testing containing 29 images. All results reported in this paper, excluding our benchmark submission, are referent to these 29 images. The evaluations are performed in the same way as the benchmark server, i.e. the prediction and the ground truth images are both converted into BEV space and are compared pixel-wise. 

We implemented our system using the Python-based SciPy software ecosystem and scikit-image library for feature extraction. We use the MLP GPU implementation provided by the Pylearn2 \cite{pylearn2_arxiv_2013} library and conduct the PSO hyperparameter optimization using the Optunity \cite{Claesen:2014} library. The tests were conducted on a machine equipped with an Intel Core i7-4930K, 64GB RAM and an NVIDIA Titan X. The GPU was utilized only for model training and testing, the rest of the system runs on a single core.

For every test and the benchmark submission, we fixed the blocks size at $10\times10$ for the classification blocks and $20\times20$ for the contextual blocks, hence we always use the additional support block. We believe that those sizes yield a good compromise of computational cost, discriminative power and classification granularity. In this work, we choose to focus on evaluating the blocks scheme itself rather than parameters effects. 

\begin{table}
\caption{Hyperparameter Search Setup}
\centering
\begin{tabular}{@{}ll@{}}
\toprule
Parameter        & Value        \\ \midrule
N. Iterations   & $10$           \\
Particles        & $10$           \\
N. Hidden Neurons & $(16, 2000)$   \\
Learning Rate    & $(0.001, 0.5)$ \\
Max. Norm Hidden & $(0.5, 5)$     \\
Max. Norm Output & $(0.5, 5)$           \\ \bottomrule
\end{tabular}
\label{tab:hpparams}
\end{table}

\subsection{Training Scheme}

To generate the features vectors (samples) for training, we used only classification blocks whose ground truth pixels are all of the same class, excluding, therefore, ambiguous cases. We adequately pad images to accommodate the selected block sizes and contextual blocks in order for the classification blocks to cover the full original image. As the top 150 lines of every image contain only negative examples and are not considered in the BEV space evaluation, we ignore this region when generating the feature vectors for training. This measure reduces the training time and helps to improve the class balance.

The samples extracted from the 260 images that are selected for training/validation are randomly split into $70\%$ training and $30\%$ validation. Each of these datasets are further subsampled at $20\%$ for the hyperparameters search, where the validation set is used for early stopping and for the hyperparameter selection. Once the best hyperparameters are established, the training proceeds by using the initial $70-30$ split, where the validation set is used only for early stopping. All samples are standardized feature-wise based on the training dataset. The hyperparameter search configuration is presented in Table \ref{tab:hpparams}. We use $10$ particles and $10$ iterations resulting in $100$ training procedures. Further MLP parameters are the $100$ mini-batch size, $0.9$ momentum and $30$ patience.

\subsection{Evaluation of Contextual and Road Blocks}

\begin{figure}
        \centering
        \begin{subfigure}[b]{0.48\textwidth}
                \includegraphics[width=\textwidth]{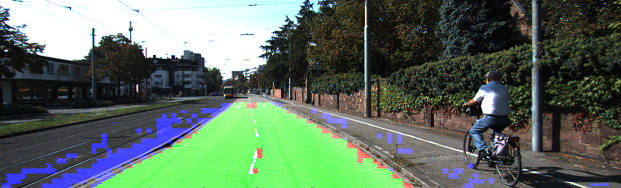}
                \caption{No contextual blocks (radius 0).}
                \label{fig:gull}
        \end{subfigure}
        
        \begin{subfigure}[b]{0.48\textwidth}
                \includegraphics[width=\textwidth]{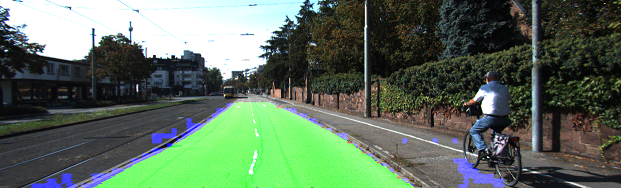}
                \caption{Radius 1.}
                \label{fig:tiger}
        \end{subfigure}
        
        \begin{subfigure}[b]{0.48\textwidth}
                \includegraphics[width=\textwidth]{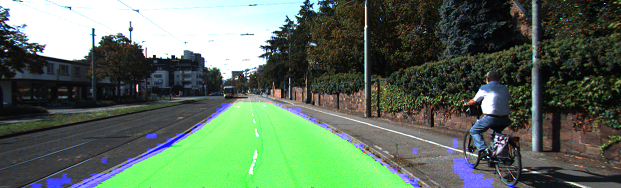}
                \caption{Radius 2.}
                \label{fig:tiger}
        \end{subfigure}
        
        \begin{subfigure}[b]{0.48\textwidth}
                \includegraphics[width=\textwidth]{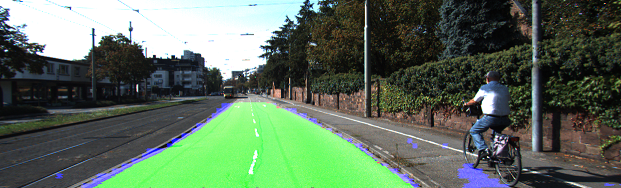}
                \caption{Radius 3.}
                \label{fig:tiger}
        \end{subfigure}
        \caption{Classification results using different radius parameter values where green represents true positive, red false negative and blue false positive.}\label{fig:evalradius}
\end{figure}

We initially evaluate the effects of using contextual blocks and their radius parameter using all image features. Table \ref{tab:radius} shows the results when varying the radius on the 29 testing images. A radius of $0$ means that no contextual block is in use. The results show a substantial increase in the F-measure from no contextual block use (radius $0$) to radius $1$ and further radius increases yield a small but consistent improvement. This effect is also clearly visible in the classification results as shown in Fig. \ref{fig:evalradius}. The image classified using no contextual blocks presents a significant amount of false positive and false negative pixels. With a radius of 1, all false negative pixels are removed and the number of false positives is reduced. The number of false negatives continues to decrease until radius 3, when the left side of the resulting image is almost clear of false positives.

\begin{table}
\caption{Contextual Blocks Radius Evaluation (in \%)}
\centering
\begin{tabular}{@{}c|cccc@{}}
\toprule
Radius & F-measure & Accuracy & Precision & Recall \\ \midrule
0      & 83.7      & 87.8     & 87.8      & 79.9   \\
1      & 86.3      & 89.4     & 87.5      & 85.0   \\
2      & 87.3      & 90.4     & 89.6      & 85.2   \\
3      & 88.2      & 91.0     & 90.2      & 86.2   \\ \bottomrule
\end{tabular}
\label{tab:radius}
\end{table}

These results highlight the validity of our contextual blocks approach and despite the higher dimensionality of the feature vector, the classifier was able to take advantage of the additional information. We did not test radiuses larger than $3$ due to hardware constraints (especially the working memory), nevertheless the benefit of larger radiuses is expected to fade and not compensate the additional computational cost.

We also evaluate how the road blocks affect the performance. For that purpose, we removed the road blocks features while maintaining the best radius parameter previous obtained ($3$) and all image features. Table \ref{tab:road} shows the results where the column ``Diff.'' refers to the difference in F-measure when using all blocks. The removal of the road blocks has a minor effect on the performance, affecting it less than a single decrease in the contextual blocks radius. We can therefore conclude that, for this dataset, our method is robust and does not depend on the usage of road blocks. However in datasets where the change in road appearance between training and test sets is more enunciated, these blocks could play a major role in helping with generalization. 

\begin{table}
\caption{Road Block Evaluation (in \%)}
\centering
\begin{tabular}{@{}c|cccc|c@{}}
\toprule
Blocks & F-measure & Accuracy & Precision & Recall & Diff. \\ \midrule
All      & 88.2      & 91.0     & 90.2      & 86.2  & 0.0\\
No Road      & 87.9      & 90.8     & 90.1      & 85.9 & -0.3 \\ \bottomrule
\end{tabular}
\label{tab:road}
\end{table}

\subsection{Features Evaluation}

Using the best radius deducted from previous experiments, we evaluated the contribution of each feature subset. To do so, we removed each feature subset and evaluated the performance on the 29 test images. The results are show in Table \ref{tab:featureseval}. These results show that the LBP and LM 2 texture features provided the most significant contribution despite LBP using the unusual 4 neighbor parameter and the small subset of filter selected for the LM features. The LM 1 features did not provide benefit and, in fact, their removal resulted in a $0.3$ F-measure increase. Considering that a non-linear parametric model is employed, we suspect that the unique information content of the LM1 features did not compensate for their relative high dimensionality ($750$ considering all blocks). The RGB features provided a reasonable contribution while the gray features made little difference, probably due to their redundancy with the RGB ones. The spatial prior showed of little importance for our method, which is expected since we use a large contextual support. Methods with smaller or no contextual support would greatly benefit from using a spatial prior. Overall, the method is robust to the feature selection as no feature subset removal reduced the F-measure to the level of not using contextual blocks.

\subsection{Processing Time}

Table \ref{tab:proctime} shows the average processing time to classify an image relating it to each major stage in our system and the contextual blocks radius parameter. These results were produced using all but LM 1 features. The feature extraction phase refers to the generation of grayscale, entropy, LBP and filtered images. The pre-processing and concatenation phase encompasses the mean, standard deviation, histogram calculations and also the feature concatenation of all blocks involved. Finally, the model prediction phase refers to the prediction time of the model for all classification blocks.

\begin{table}
\caption{Features Evaluation (in \%)}
\centering
\begin{tabular}{@{}l|cccc|c@{}}
\toprule
Feature Subset & F-measure & Accuracy & Precision & Recall & Diff. \\ \midrule
All      & 88.2      & 91.0     & 90.2      & 86.2  & 0.0\\
No RGB      & 87.7      & 90.7     & 90.4      & 85.2 & -0.5 \\
No Gray      & 88.1      & 90.9     & 90.4      & 85.8 & -0.1 \\
No Entropy      & 87.7      & 90.7     & 90.3      & 85.3 & -0.5 \\
No LBP      & 87.3      & 90.4     & 90.2      & 84.6  & -0.9 \\
No LM 1      & 88.5      & 91.2     & 90.4      & 86.8 & +0.3 \\
No LM 2      & 87.0      & 90.0     & 88.6      & 85.5 & -1.2 \\
No Spatial      & 88.0      & 90.8     & 90.4      & 85.7 & -0.2 \\ \bottomrule
\end{tabular}
\label{tab:featureseval}
\end{table}

Overall, the processing time of our system implementation is far from achieving real-time but we believe that an optimized implementation, e.g. using C language and taking advantage of multi-core systems, may achieve it. This belief is motivated by the fact that there is no stage in our system that is intrinsically costly (e.g. Textons or HoG features) and the most time consuming parts of our system (convolutions, windowed operations) are suitable for parallelization. One important thing to notice is how the processing time scales with the radius, although the number of blocks (and features) greatly increases with larger radiuses, the processing times are less affected. This is due to the efficient implementation of the contextual blocks, where their features are pre-calculated for the whole image and then appropriately concatenated for each classification block. The feature extraction phase takes longer due to the larger padding used.

\begin{table}[b]
\caption{Average Processing Time for Classifying an Image (in seconds).}
\centering
\begin{tabular}{@{}l|llll@{}}
\toprule
 Radius                               & 0     & 1     & 2     & 3    \\ \midrule
Feature Extraction                    & 0.44   & 0.49  & 0.51  & 0.65 \\ 
Pre-processing and Concatenation      & 0.94  & 1.07  & 1.17  & 1.27   \\
Model Prediction                      & 0.02  & 0.03  & 0.04  & 0.05 \\ \midrule
Total                                 & 1.40 & 1.59 & 1.72 &  1.97    \\ \bottomrule
\end{tabular}
\label{tab:proctime}
\end{table}

\subsection{Benchmark Submission}

To compare our method with others, we submitted our method results to the road detection KITTI Benchmark\footnote{\url{http://www.cvlibs.net/datasets/kitti/index.php}} using all but the LM 1 features and a radius of 3 for the contextual blocks (the best configuration according to the performed experiments). Table \ref{tab:benchmark} presents the first nine benchmark results in the \emph{Urban Road} category which includes all road images types (UU, UM and UMM). Our method achieved the fifth best score out of 31 participants, including the ones taking advantage of LIDAR (FusedCRF and RES3D-Velo) or stereo vision (NNP and ProbBoost). 


\begin{table}
\caption{Urban Road KITIT Benchmark Results (in \%)}
\centering
\resizebox{\linewidth}{!}{%
\begin{tabular}{@{}c|ccccc|c@{}}
\toprule
Method & MaxF    & Pre.   & Rec.   & FPR   & FNR & Runtime \\ \midrule
DNN \cite{Mohan:2014}    & 93.43		&95.09	&91.82	&2.61	&8.18 & 2s\\
HIM \cite{Munoz:2010}    & 90.64     & 91.62 & 89.68 & 4.52& 10.32 & 7s \\
NNP                      & 89.68    & 89.67& 89.68& 5.69& 10.32 & 5s \\ 
NED                      & 89.12  &85.80 &92.71 & 8.45 &7.29 & 1s\\
\textbf{Our method}   & 88.97    &89.50  &88.44  &5.71  &11.56 & 2s\\ 
FusedCRF   & 88.25   &83.62  &93.44  &10.08  &6.56  & 2s\\ 
ProbBoost \cite{vitor2014probabilistic}    & 87.78  &86.59  &89.01  &7.60  &10.99  & 150s\\ 
SPRAY \cite{Passani:2014}    & 87.09  &87.10  &87.08  &7.10  &12.92 & 0.04s\\ 
RES3D-Velo \cite{Shinzato:2014}    & 86.58  &82.63  &90.92  &10.53  &9.08 & 0.36s\\ \bottomrule
\end{tabular}}
\label{tab:benchmark}
\end{table}

The first two methods (DNN and HIM) uses global context (takes the whole image into consideration) which may explain their high scores. The next two methods are yet to be referenced, the only information available tells that the NNP method uses stereo vision (plane fitting) and NED uses some form of CNN. The fastest method in the benchmark is the SPRAY method. As our work, this method focuses on providing the classifier contextual cues in an efficient way. All methods scoring better than ours uses some form of parallel processing and could not achieve real-time.

\begin{figure*}
        \centering
        \begin{subfigure}[b]{\textwidth}
                \includegraphics[width=\textwidth]{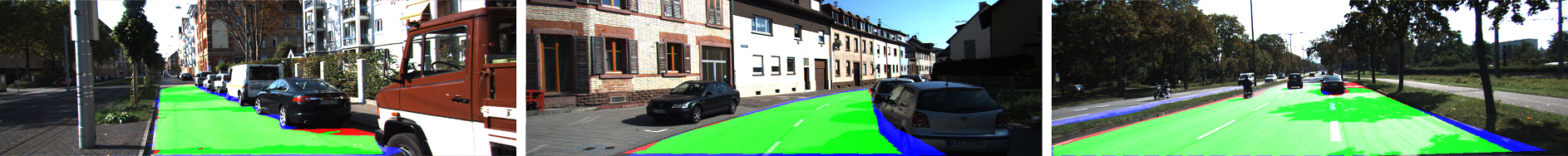}
                \caption{DNN method sample results.}
                \label{fig:gull}
        \end{subfigure}
        
        \begin{subfigure}[b]{\textwidth}
                \includegraphics[width=\textwidth]{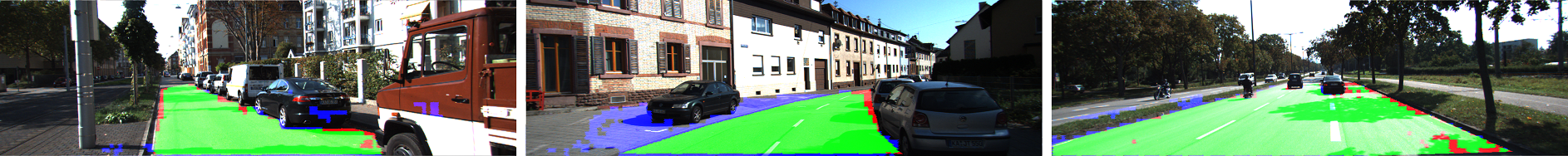}
                \caption{Our method sample results.}
                \label{fig:tiger}
        \end{subfigure}
        
        \begin{subfigure}[b]{\textwidth}
                \includegraphics[width=\textwidth]{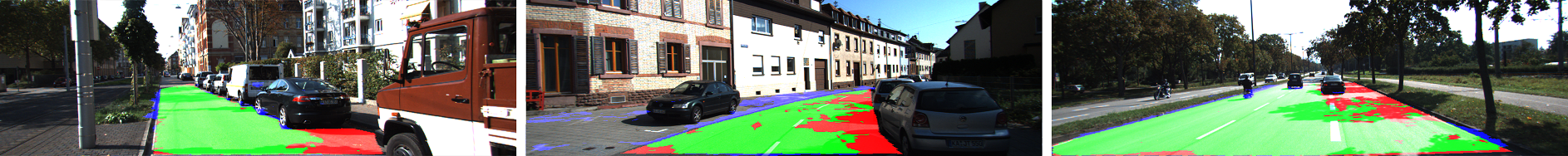}
                \caption{SPRAY method sample results.}
                \label{fig:tiger}
        \end{subfigure}
        \caption{Sample classification results extracted from the KITTI benchmark server where green represents true positive, red false negative and blue false positive.}\label{fig:kitcomp}
\end{figure*}

Figure \ref{fig:kitcomp} shows a visual comparison\footnote{Video demo: \url{http://youtu.be/QFmOZyqtClU}} of the first method, our method and the fastest one using images provided by the benchmark server. The DNN method tends to obtain smoother boundaries and an overall better result. Our method and the SPRAY one have a tendency to misclassify similar regions, but our does so to a lesser extent. Our method, however, presents a few more false positives predictions than the other two.

\section{Discussion}
\label{sec:discussion}

The proposed approach yields results in line with state of the art methods. The use of contextual blocks provides significant performance improvements that scales adequately with the radius parameter. The method run-time depends mostly on the images features selection, while the block scheme itself have a low computational cost since their features can be pre-calculated and simply concatenated afterwards. One advantage of our method is its simplicity, especially when compared to other road detection works (e.g. \cite{vitor2014probabilistic, Passani:2014}). We provide a small image features selection that seems to be adequate for road detection and whose implementation can be highly optimized. We also presented other details such as the training scheme and hyperparameters search that may have contributed to the method performance.

Despite encouraging results, our method has some limitations. Global features are unpractical to include due to padding requirements and, even with a large radius, the whole image can not be considered. The presented implementation is not optimized and, although we hold that it could be optimized for real-time purposes, we do not provide evidence that it is the case. Compared to deep learning methods \cite{Mohan:2014}, our method has the disadvantage of requiring a selection of hand-crafted features, which is mostly intuitive since it is not possible to evaluate all combinations of image features present in the literature. Finally, the use of road blocks is controversial as it is based on the assumption that the bottom part of the image always refers to a road region. In this work, however, the road blocks could be removed with a minimal performance penalty or, more generally, other sensors (e.g. stereo camera) could be used to support that assumption.

\section{Conclusions and Future Works}
\label{sec:conclusions}

This work proposed an efficient block scheme to exploit contextual information and also sensible choices for image features and classifier. Each system component has been evaluated, along with image feature subsets and processing times. The results reaffirm the importance of contextual information for road detection and demonstrate the method effectiveness that, despite being simple, could achieve results comparable with state of the art methods. Unfortunately the method still has some limitations that need to be addressed, as its inability to incorporate fully global contexts and the current implementation run time. As future work, we intend to perform an optimized implementation of our method and use convolution neural networks as feature extractors.

\section*{Acknowledgment}
The authors would like to acknowledge the support granted by FAPESP (process nr. 2011/21483-4) and CNPq (process nr. 202415/2014-7).

\bibliographystyle{IEEEtran}
\bibliography{itsc2015}


\end{document}